\pdfoutput=1

\documentclass[11pt]{article}
\usepackage[final]{coling}
\usepackage{times}
\usepackage{latexsym}
\usepackage{comment}

\usepackage{booktabs}
\usepackage[T1]{fontenc}
\usepackage{todonotes}
\usepackage{latexsym}
\usepackage{longtable}
\usepackage{float}
\usepackage{listings}
\usepackage{graphicx}
\usepackage{pdflscape}
\usepackage{lipsum}
\usepackage[utf8]{inputenc}
\usepackage{tikz-qtree}
\usepackage{forest}
\usepackage{enumitem}
\usepackage[section]{placeins}

\usepackage[title]{appendix}

\usepackage{titlesec}

\newcommand{\adam}[1]{\todo[inline,backgroundcolor=orange]{\textbf{AM:} #1}}

\newcommand{\advait}[1]{\todo[inline,backgroundcolor=yellow]{\textbf{AD:} #1}}

\titlespacing\section{0pt}{12pt plus 4pt minus 2pt}{0pt plus 2pt minus 2pt}
\titlespacing\subsection{0pt}{12pt plus 4pt minus 2pt}{0pt plus 2pt minus 2pt}
\titlespacing\subsubsection{0pt}{12pt plus 4pt minus 2pt}{0pt plus 2pt minus 2pt}

\newcommand\bfit{%
    \itshape\bfseries%
}
\usepackage{microtype}

\usepackage{soul}

\title{Semantic Role Labeling of NomBank Partitives}

\author{Adam Meyers\\
  New York University \\ \texttt{meyers@cs.nyu.edu} 
  \AND
  Advait Pravin Savant \\
   New York University \\ \texttt{as14634@nyu.edu} \\
 \\ \And
  John E. Ortega \\
  Northeastern University
\\ \texttt{j.ortega@northeastern.edu} 
   \\}

\date{}

\begin{document}
\maketitle
\begin{abstract}
This article is about Semantic Role Labeling for English partitive nouns ({\it 5\%/REL of the price/ARG1}; {\it The price/ARG1 rose 5 percent/REL}) in the NomBank annotated corpus. Several systems are described using traditional and transformer-based machine learning, as well as ensembling. Our highest scoring system achieves an F1 of 91.74\% using "gold" parses from the Penn Treebank and 91.12\% when using the Berkeley Neural parser. This research includes both classroom and experimental settings for system development.

\end{abstract}

\section{Introduction}

Semantic Role Labeling (SRL), provides a way to represent semantic concepts via labeled predicate/argument pairs. 

For example, an SRL analysis of the following three sentences include a {\bf patient} relationship between the predicate: \textit{break} and the argument \textit glass.  The 3rd sentence includes an additional {\bf agent} relationship between the predicate \textit{break} and the argument \textit{ John}.
\begin{itemize}[topsep=0pt,itemsep=-1ex,partopsep=1ex,parsep=1ex]
    \item The glass broke
    \item The glass was broken
    \item John broke the glass.
\end{itemize}

 SRL has become popular in linguistics and NLP following Jeffrey Gruber's dissertation \cite{grub:stud65}. His study focused on arguments of the main verb in a sentence ({\it They/Agent bought/Predicate a car/Theme}). Since then, the  scope of semantic role labeling has widened to cover arguments of both verbs and nonverbs. See, for example, work on PropBank, Nombank, the Penn Discourse Treebank, FrameNet among other projects \cite{Palmer-propbank,meye:nomb04,film:frame98,Miltsakaki-EtAl:2004:HLTNAACL}. Predicates can be not only verbs, but also nouns, subordinate conjunctions, adverbs or adjectives.

\noindent
\begin{figure}[hbt]
\begin{enumerate}[topsep=0pt,itemsep=-1ex,partopsep=1ex,parsep=1ex]
\item Mary's/{\bf ARG0} request/{\bf REL} for a loan/{\bf ARG1} $\rightarrow$ Mary requested a loan
\item His/{\bf ARG0} promise/{\bf REL} to improve/{\bf ARG2} $\rightarrow$ He promised to improve
\item She/{\bf ARG0} made/SUPPORT demands/{\bf REL} on her staff/{\bf ARG2} $\rightarrow$ She demanded that her staff (do something)

\end{enumerate}
\vspace{-5pt}
\caption{Sample Nominalizations}\label{nombank-nominalizations}
\vspace{-15pt}
\end{figure}

\begin{figure}[H]

\begin{description}[topsep=0pt,itemsep=-1ex,partopsep=1ex,parsep=1ex]
\item[{\bf Quant:}]
A set/{\bf REL} of tasks/{\bf ARG1}
\item[{\bf Part:}] 11 components/{\bf REL} to the index/{\bf ARG1}
\item[{\bf Meronym:}] Anthers/{\bf REL} in these plants/{\bf ARG1}
\item[{\bf Group:}] The Airline/{\bf ARG2} creditors/{\bf ARG1} committee/{\bf REL}
\item[{\bf Share:}] Her/{\bf ARG0} portion/{\bf REL} of pie/{\bf ARG1}
\end{description}
\vspace{-10pt}
\caption{Partitive Noun Examples}\label{nombank-5-examples} 
\vspace{-10pt}
\end{figure}

This article is about a semantic role labeling (SRL) task based on a subset of NomBank \cite{meye:anno04,meye:nomb04}: the identification of {\bf ARG1} arguments of predicates belonging to the five NomBank classes in Figure~\ref{nombank-5-examples}. 
Collectively, we call these {\bf partitive nouns}. In PropBank and NomBank \cite{Palmer-propbank,meye:anno04} propositions consist of a predicate (REL) and one or more aguments (ARG0, ARG1, ARG2, ... ) of  that predicate.  
13K out of the 115K NomBank propositions are partitive examples. There are a total of about 8K different noun predicates in NomBank, including about 500 partitives.  This article focuses on automatically identifying arguments of partitive nouns in the NomBank corpus. We discuss several student systems that were created during a graduate class, a baseline system, student-inspired modifications to the baseline and some ensemble systems. All development assumed gold parses. However, we also tested our final systems based on output of the Berkeley Neural parser \cite{kitaev-klein-2018-constituency}.

\section{Partitive Nouns in an SRL Framework} 

SRL can model how different sentences are nearly equivalent semantically, as illustrated in Figure~\ref{nombank-nominalizations}. Similarly, at the sentence level, the relations
{\bf ARG0(eat, clam)} and {\bf   ARG1(eat, tourists)} can be realized by the following sentences, all containing forms of {\it eat, clam,} and tourists:
\begin{itemize}[topsep=0pt,itemsep=-1ex,partopsep=1ex,parsep=1ex]
\item The giant clam ate the tourists.
\item The tourists were eaten by the giant clam.
\item The tourists, who were eaten by the the giant clam, were very wealthy.
\item The giant clam, after eating  tourists, left town.
\item The giant clam wanted to eat the tourists.
\end{itemize}

The particular labels used (like {\bf ARGO} and {\bf ARG1}) vary between frameworks, although we adopt the nomenclature of PropBank and NomBank. 

We mark the nominal predicate with the label {\bf REL} and arguments with argument labels like {\bf ARG1}, placed at the end of the phrase, e.g., in Figure~\ref{nombank-5-examples}, example 1, there is a {\bf PART} relation between the predicate {\bfit components} and the argument {\bfit the index}.  In the example, the {\bf REL} and {\bf ARG1} labels mark the predicate and the argument (via the head words). Equivalently, we can say that the relation: {\bf Part(component,index)} holds. We  also ignore the question of how the heads or extents of phrases are chosen.
We will follow a convention in which the argument is referred to by its label, i.e., if a relation {\it ARG0(eat,clam)} holds, we can refer to {\it the clam} as the {\bf ARG0} of {\it eat}.

Partitive noun predicates quantify over their {\bf ARG1} so that the relation represents a group of {\bf ARG1}s or a part of an {\bf ARG1}.  Subclasses include: {\bf quant}, {\bf part}, {\bf meronym}, {\bf group}, and {\bf share}, as in Figure~\ref{nombank-5-examples}.  Partitive nouns have several features in common. Partitive nouns tend to be transparent: this means that the {\bf ARG1} tends to act as a "semantic" head. For example, {\it a variety of sandwiches} is interpreted as an instance of {\it sandwiches} and thus edible, rather than an instance of {\it variety} and thus too abstract to eat. Coordinate conjunctions (CCs) like {\it and} and {\it or} are similar in this respect -- the conjuncts, not the CCs are semantic heads. Thus the phrase {\it the sandwich and the apple} is an instance of {\it foods} (a generalization of the two conjuncts), not the coordinate conjunction {\it and} (the word that links the other words in the phrase together).

This paper is about several SRL systems for detecting ARG1s of partitive nouns.

\section{Previous Work}

\subsection{Previous SRL Annotation}
Some previous work using NomBank SRL \cite{surd:conl08, CoNLL-2009-ST}
focuses on nominalizations (nouns related to verbs as in Figure~\ref{nombank-nominalizations}) as part of a larger task that includes verbs and their arguments). 
Other work \cite{jian:sema06} uses names of frames for non-nominalizations as features for machine learning. Our approach partitions annotation by classes of nouns that are not related to verb-frames. This approach is similar to FrameNet \cite{coll:berk98}. Arguments of nouns sharing the same frame  are treated like separate tasks.
For example, FrameNet's {\bf contingency} frame is shared by verbs, nouns and adjectives in the following examples:
\begin{itemize}[topsep=0pt,itemsep=-1ex,partopsep=1ex,parsep=1ex]
\item Success/OUTCOME may depend/VB on available resources/DETERMINANT
\item Success/OUTCOME seems dependant/JJ on available resources/DETERMINANT
\item Success/OUTCOME will be a function/NN of the Available resources/DETERMINANT
\end{itemize}
FrameNet Systems attempt to find sets of predicates and arguments belonging to particular frames. Shared Tasks such as ACE \cite{doddington-etal-2004-automatic} also follow this approach.
Frame-like classes in NomBank are noun-centric. Figure~\ref{nombank-5-examples} provides examples of the five partitive classes. In addition, there are 12 other non-nominalization frame-like classes in the NomBank dataset. (See figure~\ref{nombank-12-examples} in Appendix \ref{More-Other-NomBank-Classes}. We leave  these to future work.) This paper provides a framework for such investigations. 






In the NomBank \cite{meye:anno04} project, approximately 115,000 noun predicates were annotated, along with their arguments. The annotation scheme is compatible with the previous PropBank \cite{Palmer-propbank} annotation of verbs and their arguments. Most subsequent automatic SRL systems \cite{jian:sema06} treat NomBank propositions in a similar manner as PropBank. They use frames associated with each predicate, but they do not generalize over shared frames. For example, ARG1s of all 8000 distinct partitive predicate lemmas are assumed to form 8000 distinct classes, whereas our approach generalizes ARG1s of these 8000 predicate lemmas to a single class.

\subsection{Previous SRL Systems}



Most SRL systems fall into two 
categories. One approach is to develop a feature space from the input text based on linguistic analysis, and then extrapolate patterns via predictive modeling from feature space to output space. In the second approach, a deep learning system is created based on distributed word representations and feature spaces learned from parameterized transformations \cite{Goodfellow-et-al-2016}. These systems mainly identify relations between verb predicates and noun arguments. Figure \ref{NN-system-out} summarizes the results of several neural network systems on the CONLL 2005 \cite{carreras-marquez-2005-introduction} WSJ and Brown datasets. \\


\begin{figure}
\footnotesize
\begin{center}
\begin{tabular}{|l|l|l|}\hline
{\bf Citation}  & {\bf Detail} & {\bf F1}\\\hline
\cite{collobert2011natural}  & CNN & 77.92\\\hline
\cite{zhou2015end} & stacked LSTMs & 81.07\\\hline
\cite{he2017deep} & BiLSTM & 83.2 \\\hline
\cite{ouchi-etal-2018-span}  & span-based BiLSTM & 87.0 \\\hline 
\end{tabular}
\caption{Neural Network SRLs for CONLL 2005 data}\label{NN-system-out}
\vspace{-20pt}
\end{center}
\end{figure}
\normalsize

\cite{pradhan-etal-2005-semantic} and \cite{xue-palmer-2003-annotating} report results on PropBank \cite{kingsbury-palmer-2002-from} data. \cite{xue-palmer-2004-calibrating} experiment with various linguistic features, using representations of syntactic parse trees for use in SRL (F score 88.51 with gold parses, F score 76.21 with  \cite{coll:head99} parses). \cite{pradhan-etal-2005-semantic} develop an SRL system, integrating different syntactic views, based on experiments with CCG parsers and Charniak parsers (F score 89.4 with gold parses, F score 79.3 with  automatic parses).

 \cite{jian:sema06} work with Nombank data. They develop Nombank specific features in view of linguistic considerations, incorporate parse tree based features. A maximum entropy classification model is trained based on the generated features. For every Nombank class, there is a feature which is used denote the class.  For every nominal predicate, for corresponding words in the sentence, the class of the predicate is incorporated in the feature space. The class of a nominal predicate is
seen to be indicative of the role of its arguments. During testing, in order to enforce non-overlapping arguments, an algorithm maximizes the log probability of the labeled parse tree (F score 72.73 with gold parses, F score 69.14 with automatic parses). 


\section{Research in a Classroom Environment}
We used \% and partitive tasks as assignments for three graduate NLP classes (Spring 2012,  Spring 2022 and Fall 2023). The \% task was assigned as one of several shared-task homework assignment during the first half of the semester. During the second half of the semester the partitive task was a final project task, the basis of a final paper and final presentation. The Fall 2023 version of the task is the basis of this article. Students were permitted to use any methods they wanted for the task. For the partitive task, they were free to use our baseline system as a starting point (results are in table~\ref{baseline-results-table} and the {\bf feature based 1} rows of table~\ref{test-results-table}. Our final results include the incorporation of aspects of student systems, as discussed below.
\section{Data and Evaluation Methodology}\label{Data Section}

\subsection{Data Format}
\begin{figure}
\footnotesize
\begin{center}
\begin{tabular}{|l|l|l|l|l|l|}\hline
{\bf WORD} & {\bf POS} & {\bf BIO}& {\bf \#}
& {\bf FUNC} & {\bf FRAME}\\\hline
Output & NN & B-NP & 0 
& ARG1 &\\
in & IN & B-PP & 1 

& & \\
August & NNP & B-NP & 2 
& & \\
rose & VBD & O & 3 
& SUP & \\
5 & CD & B-NP & 4 
& & \\
percent & NN & I-NP & 5 
& PRED & QUANT \\
. & . & O & 6
& 
& \\\hline
\end{tabular}
\caption{Annotation of {\it Output in August rose 5 percent.}}\label{Annotated_Sentence}
\end{center}
\vspace{-20pt}
\normalsize
\end{figure}

For each annotated NomBank predicate in
a subset of the Penn Treebank 2 WSJ corpus (PTB), we created a representation of the sentence containing that word. The representation includes information from both PTB and NomBank. We created datasets for various classes of NomBank predicates.  In this paper, we will focus on systems trained and tested on the subset of NomBank devoted to partitive nouns, leaving NomBank subsets for nouns with other frames for future work. 

 We represent each sentence containing a partitive noun predicate as a set of tuples, one tuple for each token in the sentence. Each tuple is on one line in the data file, with blank lines between sentences. This format is an extension of the CONLL 2000 format \cite{tjong-kim-sang-buchholz-2000-introduction}.

We eliminated problematic, but rare types of examples from the data.  Since it is rare for a sentence to contain two partitive nouns (less than 1\% of all sentences with partitives), we only kept the first instance of the selected predicate type (partitive) in any given sentence. We also eliminate rare cases in which NomBank's tokenization is at odds with the PENN Treebank's tokenization, e.g.,  the single PTB token {\bf warehouse-club} is divided into two tokens in NomBank: {\bf club} is a {\bf group} noun (a subtype of partitive) and {\bf warehouse} is its ARG1. 

Figure~\ref{Annotated_Sentence} represents  a sentence containing   the word {\bf percent} and its {\bf ARG1}  {\bf Output}. Each token has the following fields: {\bf WORD}, {\bf part of speech (POS)}, {\bf CHUNK Begin-Inside-Outside (BIO) tag}, {\bf TOKEN number}, 
{\bf PRED/ARG label} and {\bf FRAME label}.

 We assume two versions of each tasks. For the ``gold'' version,  the system starts with parses and POS tags from the Penn Treebank. For the ''non-gold'' version, we use parses and POS tasks from the Berkeley Neural parser \cite{kitaev-klein-2018-constituency}. Both tasks use chunk tags, generated as per \cite{tjong-kim-sang-buchholz-2000-introduction}, the predicate noun and its classes from Nombank, and support verbs, as per NomBank. We represent NPs as their head nouns. We represent PP ARG1s as the head of the PP object, e.g., in {\bf 5 portions  of that food}, the word {\bfit food} will be 
recorded as the ARG1 of the {\bfit group} noun {\bfit portion}, not the preposition {\bfit of}. 
The "gold" versions test the SRL systems independently of parsing error, where as the non-gold  systems provide a realistic testbed for SRL performance from raw text. Both types of systems exist for much previous work in SRL.


 We initially assume that predicting the ARG1 may be a function of: properties of the head noun argument; properties of the predicate nominal; properties of the support verb (if present) and properties of paths between key items (token-based paths and parser-based paths). The latter should include paths between the nominal predicate and ARG1 noun, as well as paths between the support verbs and ARG1 nouns. The sample feature set  for our baseline system (section~\ref{baseline-sect}) takes this approach. 

 Figure~\ref{Annotated_Sentence} simplifies the SRL task in some respects. The simplification make is easy to test machine learning (M) algorithms and different sets of features, with the goal of predicting which nouns (heads of noun phrases) are ARG1s of predicate partitive nouns. As noted, the  gold data makes the task somewhat artificial, presumably resulting in higher scores than would be possible with similar, but automatically obtained features.


\subsection{Organization of Datasets}

In the Penn Treebank, the \% sign is a noun (just like the word "percent"). It is a partitive and it occurs with an ARG1 nearly 3,000 times, making it the most frequent nominal predicate in the corpus. These instances of \% are of course part of the approximately 13,000 partitive instances, the most frequent shared frame in NomBank.

 For our experiments, we created two datasets, one consisting of \% instances and a larger one consisting of partitives. We divided each set into three subcorpora: training data consists of instances from PTB directories 02 to 21; development data comes from directory 24 and test data comes from 23 (a standard data split for this corpus). The training corpus contained 1K partitive ARG1s and 2225 \% ARG1s; Dev contained 370 and 87; Test contained 550 and 150.

 We would expect most systems to do better on the \% subcorpus than the partitive subcorpus, because the former is based on arguments of the same predicate, while the latter is based on a class of predicates with some elements of meaning in common. Thus the students first made systems for the \% task and then for the partitive task.

 For initial development, we evaluate systems only on gold versions of tasks (manual parses). However, for our final results, we evaluate on both gold and non-gold versions (parser output).

\subsection{Evaluation Method}

We report precision, recall and f-scores of ARG1 in tables~\ref{baseline-results-table} and~\ref{test-results-table}. 
There is exactly one ARG1 for each sentence, as per the data format. For example, in Figure~\ref{Annotated_Sentence}, {\it Output} is labeled ARG1.For most cases, we choose the head noun of the ARG1, e.g., if the ARG1 phrase is {\bf The big prize}, a correct match requires that the system identify {\bf prize} as the ARG1. For proper noun phrases, e.g., {\bf Exxon Mobil Corp.}, we assume that any of the name words (NNP) is  correct (Exon or Mobil or Corp.).


\section{Baseline System}\label{baseline-sect}

We created a baseline system as both a proof of concept and a starting point for student systems. Through error analysis of the baseline system, the class may better understand the task. Furthermore, it is possible that an improved system could be created both by modifying the baseline system based on features from student systems and/or ensembling it with student systems.
 First we used Sklearn's Adaboost machine learning algorithm \cite{Freund:1995:DTG},  with features generated for each token in each sentence. The model was generated from our training corpus 
 and it was used to predict the ARG1 (true or false) status of tokens in the development and test corpora. All experimentation with different features was measured against the development set.  Prior to submission of this paper, we ran one system on the test corpus, the one that performed best on the development set. 

\subsection{Baseline Features and their Motivations}\label{baseline-feature-sect}

In this section, we enumerate the features used in the baseline system. These features are motivated by our linguistic understanding of SRL with noun predicates in general, and partitives in particular. This includes both: information that characterizes typical ARG1s and predicate nouns (the head words, POS classes, BIO tags, nearby words, frame classes of predicate nouns, etc.) and information about how ARG1s are related to the predicates (token distance, tree distance, etc.). The baseline system provides a testing ground for such features.

\subsubsection{Simple ARG1 Features} Simple ARG1 features include the {\bf word} (head of ARG1), its {\bf POS} tag, its  {\bf BIO} tag, and these same features for the two words following and preceding the head. Following Firth \cite{firth-syn57} and  subsequent NLP research, we are assuming that words are, in part, defined by their context.
Thus features of neighboring words are assumed to bear on meaning of the current ARG1. 

\subsubsection{Embedding-based Features} Word embeddings, generated from large corpora, are also  compatible with Firth's principle because each word is defined based on a neural network  model of the context of that word. We use pre-trained embeddings from SPACY's pre-trained en\_core\_web\_md  tok2vec embeddings which was from a medium sized (43 mb) corpus. For each example ARG1 in the training corpus, we calculate a total of 10 embeddings: two types of embeddings for each of five different n-grams. The five n-grams include: the head ARG1 word itself, and the forward and backward bigrams and trigrams. For example, for the sentence: {\it The consumer price index/{\bf ARG1} rose five percent/{\bf REL}.},  five n-grams are generated: 1. {\bf consumer price index}, 2. {\bf price index}, 3. {\bf index}, 4. {\bf index rose}, and {\bf index rose five}. For each n-gram, we calculate two embeddings: the normal embedding for that n-gram and 
and the embedding of all words in the sentence with the n-gram removed. We will refer to the second type of embedding as a {\bf slash} embedding.
For each of these 10 types of embeddings, we calculate an average embedding based on the training file. For each ARG1 in the corpus, we find how similar (cosine similarity) the 10 embeddings are to these averages. These similarities correspond to 10 feature values.

\subsubsection{Predicate Class Features} The specific partitive class of the noun, as well as any other NomBank class the predicate is labeled with, e.g., {\bf 
GROUP, MERONYM, PART, QUANT, SHARE, BOOK-CHAPTER, BORDER, CONTAINER, DIVISION, ENVIRONMENT, INSTANCE-OF-SET, NOM, NOMADJ PART-OF-BODY-FURNITURE-ETC, SHARE, WORK-OF-ART}. These labels are taken from the lexical entries of each predicate. The labels for well-defined classes (the classes exemplified in Figures~\ref{nombank-5-examples} and~\ref{nombank-12-examples}) are used consistently according to the aforementioned specifications. The  remaining predicate labels are chosen by the annotators  during the creation of NomBank. None of these features are used for the {\bf \%} task, since these values would be the same for all instances of {\bf \%}. 

\subsubsection{Path Features}\label{path-section} Simple path features are based on token distance: number of tokens between predicate and ARG1, tokens between support and ARG1. Positive and negative values reflect the ARG being before or after the support verb or nominal predicate.  

 In the initial baseline systems ({\bf Feature based 1 and 2} in Table~2, we
 approximate paths in the parse tree, based on sequences of BIO tags, collapsing BI* to phrase sequences and including the direction. For example, consider the following sequence with BIO tags from a predicate to its ARG1: 

%
\smallskip
\noindent {\bf 20\textbackslash B-NP \%\textbackslash I-NP of\textbackslash B-PP the\textbackslash B-NP pie\textbackslash I-NP}

\smallskip

\noindent
The path feature generated from this sequence is:
\smallskip

\noindent
{\bf right\_NP\_PP\_of\_NP\_NOUN}

\smallskip

This characterizes the path from a head noun to the object of a complement of the preposition {\bf of}. Thus different left and right tags characterize paths between predicates and arguments, and between support verbs and arguments. These path features are derived by essentially collapsing sequences of BI* into phrases, e.g., B-NP I-NP I-NP $\rightarrow$ NP.
 The path features encode  simulated paths as \textit{type 1 path features} in order to differentiate them from \textit{type 2 path features}. We will henceforth refer to this method as {\bf Path-heuristic 1}.
\\

For {\bf feature-based system 3}, we use a parse tree for generating paths between words. We developed {\bf Path heuristic 2}, based on  observations about sentence structure. This heuristic  represents classes of observed grammatical relationships between: the predicate noun and the (head of) ARG1; and also between a support verb and the (head of) ARG1.

{\bf Path heuristic 2} labels one of four paths for compatible \cite{manning2014stanford} parse trees:
\begin{description}
    \item [{\bf PATH1}] From the support verb to a preceding ARG1, we assume $\mathbf{\uparrow VP \uparrow S \downarrow NP}$
    \item [{\bf PATH2}] From the support verb to a following ARG1, we assume $\mathbf{\uparrow VP \downarrow NP}$
    \item [{\bf PATH3}] From a predicate noun to a preceding ARG1 (with no support verb), we assume $\mathbf{\uparrow NP \downarrow NP}$ 
    \item[{\bf PATH4}] From a predicate noun to a following ARG1. Derive the path from the sequence of BIO tags from the predicate noun to the ARG1 (similar to the path heuristic used in {\bf Feature based systems 1 and 2}). This includes the common path from predicate to ARG1: $\mathbf{\uparrow NP \downarrow PP \downarrow NP}$ ({\bf 5\% of the price.})
    \end{description}


For example, consider the sentence. "The price rose 5 percent." and the parse tree:
\begin{verbatim}
(S
  (NP (DT The) (NN price))
  (VP (VBD rose)
    (NP (CD five) (NN percent)))))
\end{verbatim}
\noindent
Here, "percent" is the predicate, "rose" is a support verb and "price" is the ARG1. 
For words preceding the support verb "rose", we can investigate the application of PATH3, which estimates paths between a support verb and a preceding ARG1. Starting with "rose", we go up a verb phrase ($\uparrow$VP), then down a noun phrase ($\downarrow$NP).  Similarly, we could apply PATH2 in the following tree (for {\bf They increased the price five percent}): 
\begin{verbatim}
(S
  (NP (PRP They) 
  (VP (VBD increased)
      (NP (DT the) (NN price))
      (NP (CD five) (NN percent)))))
\end{verbatim}
\noindent
The support verb {\bf increased} is linked to the following ARG1 {\bf the price}, by going up to the VP and then down to the following NP.




Type 1 path features consist of encoding the actual path sub-sequence and ordinal or one-hot encoded representations. For each path feature variable, there are multiple possible path values. Type 2 path features consist of three binary features that check for the presence or absence of specific paths (our first three heuristics). 

 We implement a system with the above mentioned type 2 path features in combination with other baseline features, path features, and use a parse tree for generating paths. In comparison with the standard baseline system, we see that the incorporation of parse tree based path information and the above mentioned heuristics aids in predictive modeling for nominal SRL.

\subsection{One Hot Encoding}\label{one-hot-section}

The results in table~\ref{baseline-results-table} correspond to versions of feature based systems in table~\ref{test-results-table}. As we develop our feature based system, we consider different encoding schemes for our categorical features. Feature based system 2, uses one hot encoding over ordinal encoding used in Baseline system (feature based) 1. In table~\ref{test-results-table}, we suspect the jump in F-score from system 1 to system 2 is associated with the use of better path features by virtue of one hot encoding. The embedding based numeric features are not subject to change based on encoding schemes.
\smallskip

Consider a path feature which has n distinct values corresponding to n paths. Numbering different paths from 0 to (n-1) may not have allowed the model to yield predictive utility (ARG1 vs not-ARG1). The model has to predict the first path as 0, the second path as 1, and so on.  By representing the path features in terms of one hot encoding, the model may be able to better use the path features for downstream prediction. We see this empirically, in our F-scores. Imposition of an ordinal relationship between non-ordinal entities like parts of speech tags would imply an ontology which may not exist. Incorporating ordinal information in our model would not be the most appropriate definition of the features since it would entail us putting in additional assumptions not corroborated with linguistic phenomena. Feature based system 3 expands on Feature based system 1. It explicitly uses a parser based system via Path Heuristic 2. It uses one hot encoding rather than ordinal encoding. It also uses the features from previous systems. 
 


\subsection{Baseline Results}

\begin{table}
\begin{tabular}{|l|l|l|l|}\hline
{\bf System} & {\bf Prec} & {\bf Rec} & {\bf F1}\\\hline
\multicolumn{4}{|c|}{\bf \% Task}\\\hline

{\bf All} & {\bf 83.33} & {\bf 51.72} & {\bf 63.83} \\\hline
{\bf N-gram Only}& 60.0 & 7.24 &  26.79 \\\hline
{\bf All but Path} & 65.22 & 34.48 &  45.11\\\hline
{\bf All but Embed} & 84.09 & 42.53 &  56.49\\\hline
{\bf All but Basic Embed} & 88.37 & 43.68 &  58.46\\\hline
{\bf All but Slash Embed} & 84.31 & 49.43 &  62.32\\\hline

\multicolumn{4}{|c|}{\bf Partitive Task}\\\hline

{\bf All} & {\bf 87.08} & {\bf 48.92}  & {\bf 62.65}\\\hline
{\bf N-gram Only}  & 79.17 & 25.54 & 38.62 \\\hline
{\bf All but Path} & 77.6 & 26.08 &  39.03\\\hline
{\bf All but Embed} & 87.75 & 48.12 & 62.15 \\\hline
{\bf All but Basic Embed} & 86.05 & 49.73 & 63.03 \\\hline
{\bf All but Slash Embed} & 85.32&  50.00& 63.05 \\\hline

\end{tabular}
\caption{Baseline Results on (Gold) Development Sets}\label{baseline-results-table}
\vspace{-15pt}
\end{table}

The baseline system 1 results on the Gold task are discussed here.
As illustrated in Table~\ref{baseline-results-table}, all of the feature types contributed to the results on the development set. Thus removing a type of feature resulted in lower precision and/or recall, optimizing for an F1 score.   The development sets for {\bf \%} and {\bf partitive}  included, respectively,  87 and 372 instances of ARG1s.  The training corpus had 2225 and 9987 instances respectively. As per section~\ref{baseline-sect}, we only ran a limited set of systems on the test set, including the ALL system from table~\ref{baseline-results-table} (50 instances of {\bf \%} and 555 instances of {\bf partitives}). 

 Both embedding features (characterizing ARG1s for these classes in general) and path features (characterizing relations between ARG1 and the predicate and ARG1) contributed to the final result. Both tasks get some positive result, even if we only use the most primitive of N-gram features. 
Both types of embedding features contributed to the final results as well, indicating that both the head ARG1s (normal embeddings) and words in ARG1 contexts (slash embeddings) are characteristic of ARG1s.

\section{Student Systems}
\label{sec:student_systems}

Student systems included modified versions of the baseline system, systems with similar architectures and pure transformer-based systems, primarily using versions of BERT \cite{devlin-etal-2019-bert}. 
 In Figure~\ref{test-results-table}, we provide improved results for systems that combine student system innovations with the baseline system. The remainder of this section will be about these systems.


\subsection{Baseline System Modifications}



 Building on the baseline features, student group 16 uses a random forest \cite{breiman2001random} in order to predict ARG1s based on the same input features as the baseline. They also do a grid search across models in order to obtain an estimator. For the best performing random forest estimator, they evaluate the relative importance of features. Features representing the distance from the predicate and embedding distances are seen to have a higher predictive importance.  The f-measure increases slightly to 64.41 (percent) and  62.72 (partitive).

\smallskip

 Student Group 15's system
achieved f-measures of 82.91 (\% task) and 77.46 (partitive task). They made two significant improvements on the baseline system: they changed the embeddings to one-hot embeddings and they used better path features. They have a set path of features, derived from the sequence of BIO chunk tags between a given word and the predicate, the support verb in a sentence. 
\vspace{-5pt}

\subsection{Transformer Based Systems}






Student group 2 used a  multi-layer-perceptron classifier with BERT \cite{devlin-etal-2019-bert} embeddings, achieving 91.86 F-score for the {\bf \%}  and 79.2  for partitives. 
Student group 9 uses DistilBert \cite{sanh2019distilbert} for fine-tuning, with  tokens as features, achieving an 88.6 F-score on the partitive task. They report a 2.6 \% F-score improvement using DistilBert vs Bert. 
 Student group 10 uses a BiLSTM \cite{hochreiter1997long} with similar features as baseline system 1, plus fasttext word embeddings. They report an F-score of 78.54 for the  {\bf \%} task. 

\subsection{Deep Learning Using Linguistic Features}\label{deep-ling-sect}

Student group 4 studied the integration of linguistic features with the representational capacity of attention based deep neural networks \cite{wu-etal-2018-evaluating} \cite{sachan2021syntax}. Works such as \cite{clark-etal-2019-bert} \cite{tenney2019bert}  postulate that pre-trained large language models such as BERT implicitly learn to represent syntactic information without any inductive bias. Student group 4 considered that engineered features based on linguistic considerations could be of utility in predictive modeling for our task. This is corroborated by our baseline systems which are based on linguistics oriented feature engineering. They formulate the learning of linguistic features based on POS tags, BIO tags and directed parse tree distances of a word to the predicate as an auxiliary task performed by a separate head. This auxiliary head has a shared base representation with the BERT model which performs the downstream task of predicting semantic roles. Since the base representation is shared, the model is pushed to learn representations which can perform these tasks simultaneously, potentially enable the learning of better representations for the downstream task. 
This approach was used for {\bf deep learning system 2} in table~\ref{test-results-table} and achieved higher results than deep learning system 1.

\section{Ensembling}



Based on principles of multi-view learning and ensemble learning \cite{sun2013survey,dong2020survey}, we develop an ensemble model. Governed by linguistics, an underlying data-generating process produces sentences along with their predicates and corresponding semantic roles as input, output pairs. For every model, as we make a choice of representation to define the input and output data, we are modeling a view of the data-generating process. The performance of a model is proportional to the potency of a view to faithfully represent this process. It can be of benefit to construct a model which aggregates/processes information obtained from multiple views.

\smallskip
Our ensemble system consists of a purely deep learning component and a purely feature-based component. The deep learning component consists of a BERT-based model \cite{devlin-etal-2019-bert} that uses distributed word vector representations as input. The feature-based component consists of an Adaboost model\cite{Freund:1995:DTG} that uses the features motivated by our baseline system as input. We make predictions on semantic roles based on the aggregation of outputs of a BERT-based model and a feature-based model. We develop a voting scheme for deriving the ensemble output based on a weighted aggregation of the outputs of the two models. We learn the weights adaptively as we train the ensemble model on the data. \cite{6797059, eigen2014learning, 10.5555/1162264}. 


\section{Results Summary}\label{results_summary}

\begin{table}[t]
\begin{tabular}{|p{3cm}|p{0.9cm}|p{0.9cm}|p{0.9cm}|}\hline
{\bf System} & {\bf Prec} & {\bf Rec} & {\bf F1}\\\hline
\multicolumn{4}{|c|}{\bf \% Task}\\\hline

{\bf 1. Feature based 1} & 80.78 & 53.85 & 64.63\\\hline
{\bf 2. Feature based 2} & 93.59 & 74.51 & 82.95\\\hline
{\bf 3. Feature based 3} & 94.47 & 77.93 & 85.41\\\hline
{\bf 4. Deep learning 1} & 97.96 & 87.64 & 92.52\\\hline
{\bf 5. Deep learning 2} & 98.46 & 87.82 & 92.84\\\hline
{\bf 6. Ensemble} & 98.24 & 92.33 & 95.19 \\\hline

\multicolumn{4}{|c|}{\bf Partitive Task}\\\hline

{\bf 1. Feature based 1} & 84.58 & 48.65 & 61.76\\\hline
{\bf 2. Feature based 2} & 86.20 & 70.69 & 77.68\\\hline
{\bf 3. Feature based 3} & 91.64 & 81.15 & 86.07\\\hline
{\bf 4. Deep learning 1} & 96.55 & 82.18 & 88.79\\\hline
{\bf 5. Deep learning 2} & 97.36 & 82.41 & 89.26\\\hline
{\bf 6. Ensemble} & 98.79 & 85.62 & 91.74\\\hline

\end{tabular}
\caption{Results on Gold Test Set}\label{test-results-table}
\end{table}

\begin{table}[t]
\centering
\begin{tabular}{|p{3cm}|p{0.9cm}|p{0.9cm}|p{0.9cm}|}
\hline
{\bf System} & {\bf Prec} & {\bf Rec} & {\bf F1} \\
\hline
\multicolumn{4}{|c|}{\bf \% Task} \\
\hline
{\bf 1. Feature based 1} & 76.82 & 53.32 & 62.94 \\
\hline
{\bf 2. Feature based 2} & 86.59 & 75.76 & 80.81   \\
\hline
{\bf 3. Feature based 3} & 88.63 & 78.69 & 83.37 \\
\hline
{\bf 4. Deep learning 2} & 97.05 & 88.64 & 92.66 \\
\hline
{\bf 5. Ensemble } &  98.97 & 90.93 & 94.78 \\
\hline
\multicolumn{4}{|c|}{\bf Partitive Task} \\
\hline
{\bf 1. Feature based 1} & 50.58 & 71.74  &  59.33 \\
\hline
{\bf 2. Feature based 2} & 83.53 & 69.08 & 75.62 \\
\hline
{\bf 3. Feature based 3} & 92.34 & 76.85 & 83.89 \\
\hline
{\bf 4. Deep learning 2} & 95.59 & 83.40 & 89.08 \\
\hline
{\bf 5. Ensemble} & 98.76 & 84.58 & 91.12 \\
\hline

\end{tabular}
\caption{Results on Non-Gold Test Set}\label{test-non-gold}
\vspace{-10pt}
\end{table}

 Table~\ref{test-results-table}  lists the results on the gold test set. Table~\ref{test-non-gold} runs these same systems on the non-gold test set, providing insight to how our systems would perform on a realistic pipeline, beginning with raw text. Appendix \ref{visualizations} provides additional visual comprehension of the results. Feature based system 1 is our baseline system, with feature engineering done based on linguistic considerations. 
 Feature based system 2  modifies the baseline system to use  one hot encoding instead of ordinal encoding for feature generation. It also includes additional features denoting paths between a word in the sentence and the predicate. For example, a measure of the path distance to the predicate is considered based on the numbers of BIO chunks in between the predicate and a concerned word. One hot encoding gives concrete improvements, as initially demonstrated in a student system (Section~\ref{one-hot-section}). 
 Along with type 1 path features, Feature based system 3 incorporates type 2 path features (section~\ref{path-section}). 

 Inspired by student work, deep learning system 1 uses the BERT base \cite{devlin-etal-2019-bert} to provide contextualized word embeddings, and a fine tuned BiLSTM. Positional embeddings are significant in this set up. Concatenating predicate indicator embeddings with word embeddings after the BERT base forward pass helps the downstream deep learning model.  The tasks involve predicting if a word is an ARG1 of a PRED. Considering that each sentence has one ARG1 for a given PRED, the number of words to be classified as not ARG1 would be higher than the number of words to be classified as ARG1. Scaling the loss function for class imbalance appropriately improves the modeling process. 

\smallskip

 We evaluate linguistic features in a deep learning framework \cite{wu-etal-2018-evaluating}, deep learning system 2 modifies system 1. As discussed in section~\ref{deep-ling-sect}, apart from modeling for the downstream task of predicting ARG1, we take a weighted average of the intermediate BERT presentations, analogous to ELMo \cite{peters-etal-2018-deep}, in order to model for an auxiliary task.
The ensemble system combines the methods of the component systems and ultimately achieved the highest score.

Tables~\ref{test-results-table} and~\ref{test-non-gold} differ in a predictable pattern: systems run on the gold data are slightly better than systems run on non-gold tasks. Given parsing error is close to 5\%, this result is expected.\footnote{Deep Learning 1 is absent from Table~\ref{test-non-gold} as it does not use any gold features and thus would be the same as in Table~\ref{test-results-table}.}

 The gold data, accurately representing the data generating process, helped our models achieve better performance. Variations between gold annotations and non-gold annotations will be contingent on the properties of the parser used. When trained on non-gold data, models can adapt to these variations and learn representations that yield predictive performance, albeit slightly lower as compared to gold data. For predictive performance on a dataset, models need only differentiate between ARG1 labels and non-ARG1 labels in the created feature space. Systems on the partitive task have generally lower F-scores than those on the percent task. Additionally, for non-gold systems, those on the partitive task show a slightly larger relative decrease in F-scores. This is as expected, as the partitive task is more complex than the percent task.   

\section{Concluding Remarks}

We observed that a variety of techniques contributed to our best result: traditional ML features, deep learning with/without linguistic features and multi-view ensembling. A cooperative shared task turned out to be an excellent research model. 
We demonstrated that partitive nouns, the largest frame-like class of Nombank nouns provide a high-performing testbed for SRL, and possibly related tasks, like event and relation extraction.
   We intend to make our dataset available for research purposes.\footnote{Details will be provided on the NomBank website: \url{https://nlp.cs.nyu.edu/meyers/NomBank.html}. There may be licensing considerations because NomBank annotates  Wall Street Journal text.}  

 We may explore avenues of further research, including:  meta learning \cite{finn2017modelagnostic} across noun categories; parse tree path embeddings \cite{roth-lapata-2016-neural}; syntax infused models \cite{sachan2021syntax}; and in-context learning with large language models \cite{brown2020language}.

 \section{Limitations of Our Approach} Partitives are possibly the simplest NomBank noun class. They usually are accompanied by one argument and rarely occur with two. Their ARG1s have few semantic limitations -- they need to be measurable by the partitive predicate, e.g., {\it a pound of meat} is OK, but {\it a pound of sincerity} is semantically ill-formed. We would expect that  arguments of other NomBank noun classes (Appendix~\ref{More-Other-NomBank-Classes}) would be more difficult to classify. Systems would achieve lower scores and may require other approaches.

 Furthermore, the WSJ corpus represents the financial news register of 1990s English. Amounts in general and partitives in particular may be idiosyncratic to these data. Running our system and analyzing results for other corpora may provide further insight on semantic roles for noun predicates.

\section*{Acknowledgments} We would like to acknowledge several people who made this study possible. Ang Sun was the teaching assistant in the Spring 2012 Graduate NLP class. He was instrumental in creating the original percent and partitive task, including formatting the data and advising students. Abishek Shyamsunder, a teaching assistant in the Fall 2022 class was instrumental in organizing the student projects, distributing the baseline system and figuring out which student work we should attempt to replicate. The students in the Spring 2012, Spring 2022 and Fall 2022 built percent and partitive systems. They provided a testbed and helped us improve the task. Indeed, we based several components of our final system on work completed by students in the Fall 2022 class. This includes work by student group 2 (Tanran Zheng, Wentao Chen, Xiangyuan Wang, Xinru Xu), student group 4 (Advait Savant,	Sudharsana Kannappan Jaya Lakshmi, Yulu Qin), student group 8 (Anoushka Gupta, Arunima Mitra), student group 9 (Xiaohan Wu, Xiaoyun Zhi, Yulin Hu),  student group 10 (Prapti Devansh Trivedi, Tanya Pushkin Garg), student group 12 (Anna Xie, Yi Yan Ng, Yifan Zhang), student group 15 (Haozhen Bo, Kevin Tong, Namas Mankad), student group 16 (Jeremy Lu).

\bibliography{anthology,bibliography}


\newpage

\begin{appendices}

\section{Other Nombank Classes}\label{More-Other-NomBank-Classes}

\begin{figure}[htbp]
    \centering
    \begin{description}[topsep=0pt, itemsep=-1ex, partopsep=1ex, parsep=1ex]
        \item[{\bf ENVIRONMENT:}] A period/{\bf REL} of industrial consolidation/{\bf ARG1}
        \item[{\bf ABILITY}] Their/{\bf ARG0} right/{\bf REL} to talk/{\bf ARG1} 
        \item[{\bf WORK-OF-ART}] Congress/{\bf ARG0}’s idea/{\bf REL} of reform/{\bf ARG1} 
        \item[{\bf CRISS-CROSS}] Topics/{\bf REL} of interest/{\bf ARG1}
        \item[{\bf RELATIONAL (ACTREL)}] His/{\bf ARG2} math/{\bf ARG1} professor/{\bf ARG0}+{\bf REL} 
        \item[{\bf RELATIONAL (DEFREL)}] Mindy Hymowitz’s/{\bf ARG1} mother/{\bf ARG0}+{\bf REL}
        \item[{\bf ATTRIBUTE}] The company/{\bf ARG1} 's value of \$3 Billion/{\bf ARG2}
        \item[{\bf JOB}] the new Treasury/{\bf ARG2} post/{\bf REL} of inspector general/{\bf ARG0}
        \item[{\bf HALLMARK:}] the cornerstone/{\bf ARG0}+{\bf REL} of Phillips’ chemicals operations/{\bf ARG1}
        \item[{\bf VERSION:}] the House/{\bf ARG0} version/{\bf REL} of the deficit-cutting bill/{\bf ARG1}
        \item[{\bf TYPE:}] his brand/{\bf REL} of Christianity/{\bf ARG1}
        \item[{\bf ISSUE:}] the subject/{\bf REL} of U.S. direct investment in Japan/{\bf ARG1}
        \item[{\bf FIELD:}] the rapidly growing field/{\bf REL} of bio-analytical instrumentation/{\bf ARG1}
        \item[{\bf EVENT:}] the drought of 1988/ARGM-TMP
    \end{description}
    \caption{More NomBank "Other" Classes.}
    \label{nombank-12-examples}
\end{figure}

\vspace{1cm} 

\newpage

\section{Visualizations}\label{visualizations}


\begin{figure}[htbp]
    \centering
    \includegraphics[width=0.5\textwidth, keepaspectratio]{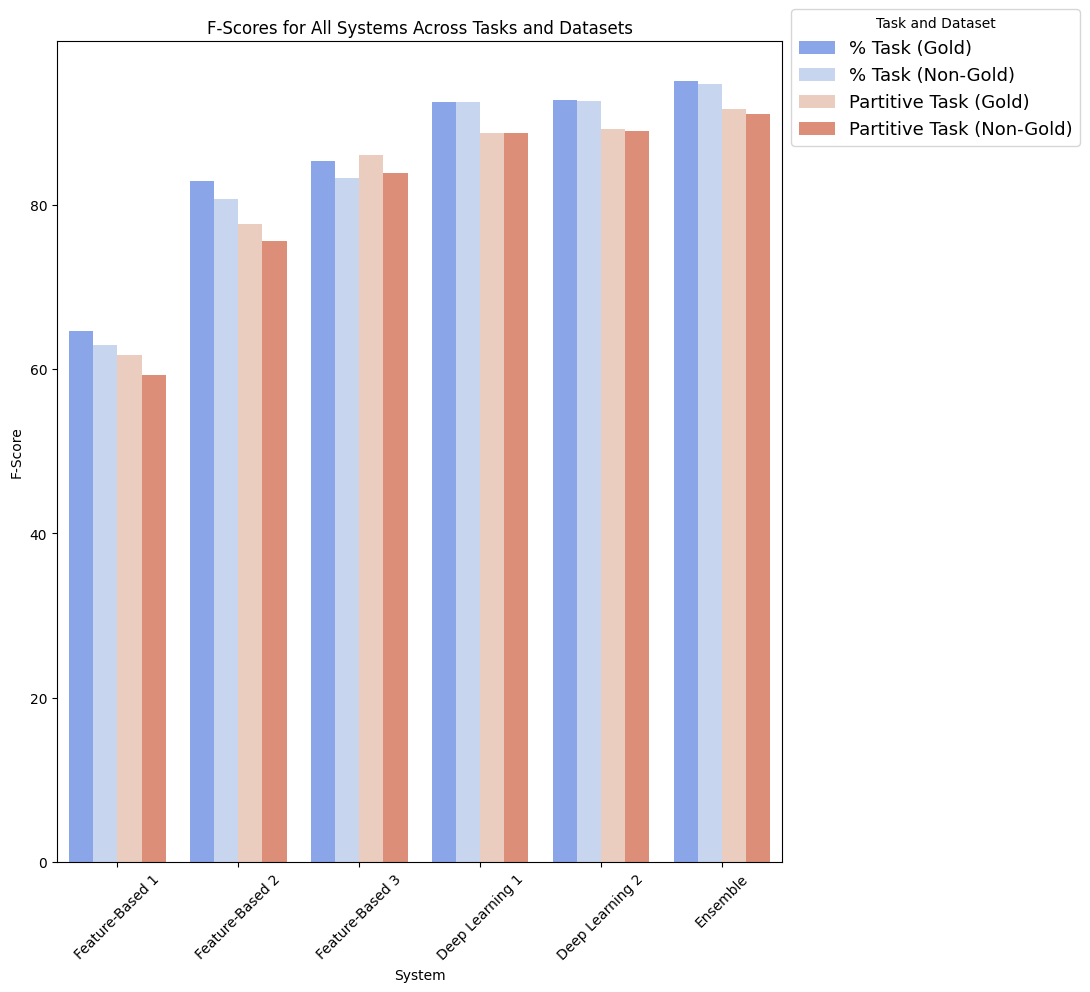}
    \caption{Bar Chart showing F-scores for different systems and datasets.}
    \label{fig:bar_chart}
\end{figure}

\vspace{1cm}

\begin{figure}[htbp]
    \centering
    \includegraphics[width=0.5\textwidth, height = 6cm, keepaspectratio]{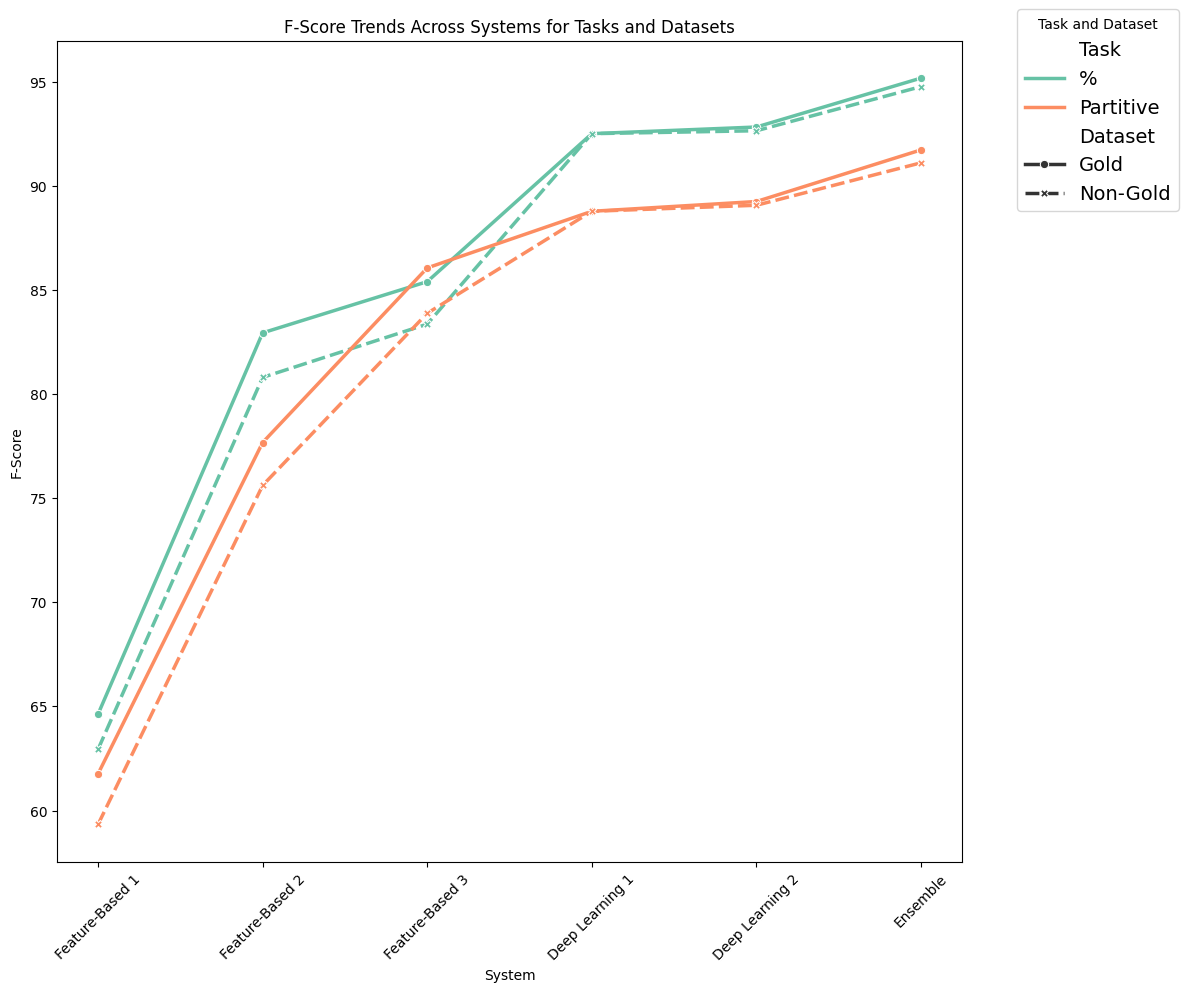}
    \caption{Line Chart displaying F-scores across systems for gold and non-gold datasets.}
    \label{fig:line_chart}
\end{figure}

\vspace{1cm} 

\begin{figure}[htbp]
    \centering
    \includegraphics[width=0.5\textwidth, height = 6cm, keepaspectratio]{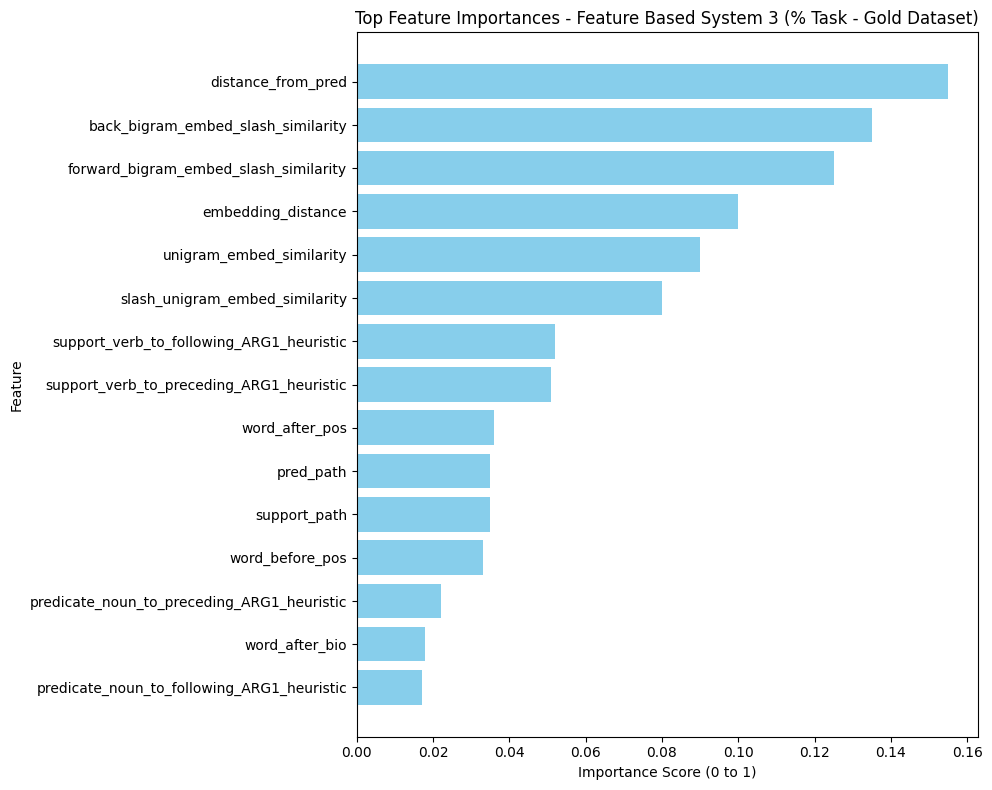}
    \caption{Feature Importance for our best feature-based system on the Percent task.}
    \label{fig:feat_imp_chart_percent}
\end{figure}

\vspace{0.5cm} 

\begin{figure}[htbp]
    \centering
    \includegraphics[width=0.5\textwidth, height = 6cm, keepaspectratio]{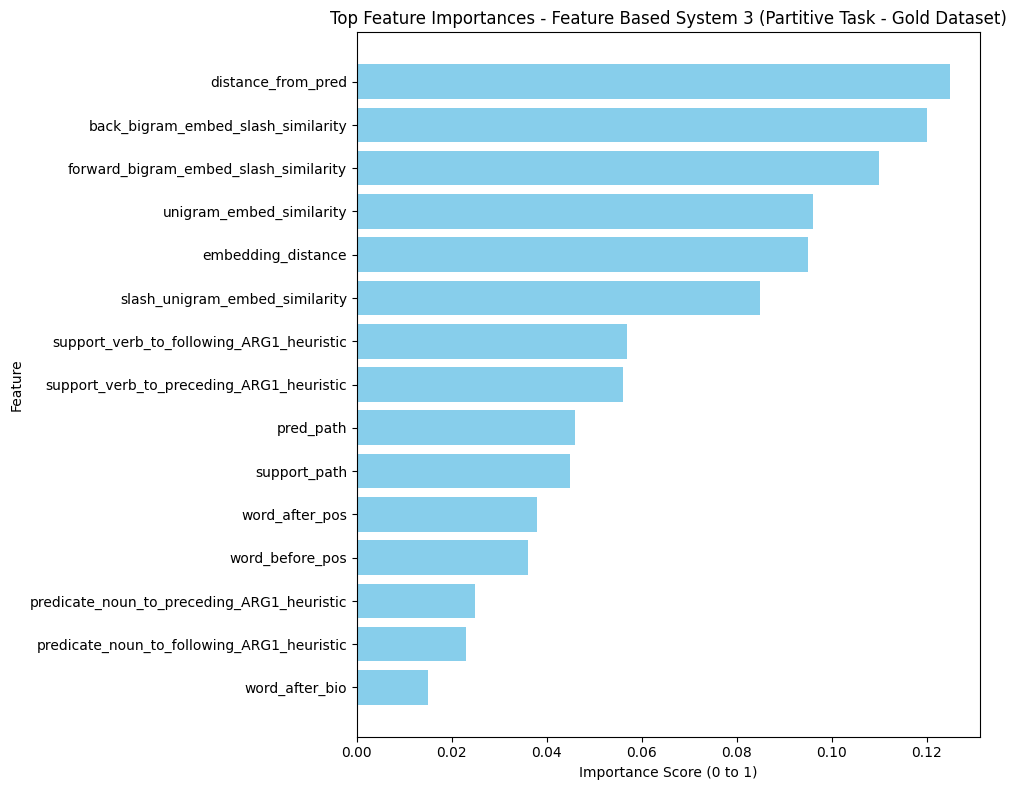}
    \caption{Feature Importance for our best feature-based system on the Partitive task.}
    \label{fig:feat_imp_chart_partitive}
\end{figure}

\newpage

\subsection{Description of figures}

Figures \ref{fig:bar_chart} and \ref{fig:line_chart} demonstrate our results and complement tables \ref{test-results-table} and \ref{test-non-gold}. In figures \ref{fig:feat_imp_chart_percent} and \ref{fig:feat_imp_chart_partitive}, the relative importance score for top 15 features (sorted in order of magnitude) is displayed. This score is derived from the Adaboost model \cite{Freund:1995:DTG} used for training feature based system 3, with descriptions as elaborated in sections \ref{results_summary}, \ref{baseline-sect}.

\end{appendices}

\end{document}